\documentclass{article}

\usepackage{arxiv}

\usepackage[utf8]{inputenc} 
\usepackage[T1]{fontenc}    
\usepackage{hyperref}       
\usepackage{url}            
\usepackage{booktabs}       
\usepackage{amsfonts}       
\usepackage{nicefrac}       
\usepackage{microtype}      
\usepackage{lipsum}		
\usepackage{graphicx}
\usepackage{natbib}
\usepackage{doi}
\usepackage{amsmath, tabularx, multirow, longtable, pdflscape, ragged2e, url, placeins}
\usepackage{algorithm}
\usepackage{algorithmic}

\title{Saying More Than They Know: A Framework for Quantifying Epistemic-Rhetorical Miscalibration in Large Language Models}

\author{ \href{https://orcid.org/0000-0002-9516-9153}{\includegraphics[scale=0.06]{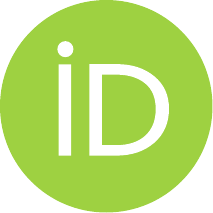}\hspace{1mm}Asim D.~Bakhshi} \\
	National University of Science and Technology\\
	Islamabad, Pakistan 46000 \\
	\texttt{asim.dilawar@mcs.edu.pk} \\
}

\renewcommand{\headeright}{A Preprint}
\renewcommand{\shorttitle}{Saying More Than They Know}

\hypersetup{
pdftitle={Saying More Than They Know: A Framework for Quantifying Epistemic-Rhetorical Miscalibration in Large Language Models},
pdfsubject={cs.AI},
pdfauthor={Asim D.~Bakhshi},
pdfkeywords={Large Language Models, AI Evaluation, Bias and Fairness, Epistemic Uncertainty},
}

\begin{document}
\maketitle

\begin{abstract}
Large language models (LLMs) exhibit systematic miscalibration with rhetorical intensity not proportionate to epistemic grounding. This study tests this hypothesis and proposes a framework for quantifying this decoupling by designing a triadic epistemic-rhetorical marker (ERM) taxonomy. The taxonomy is operationalized through composite metrics of form-meaning divergence (FMD), genuine-to-performed epistemic ratio (GPR), and rhetorical device distribution entropy (RDDE). Applied to 225 argumentative texts spanning approximately 0.6 Million tokens across human expert, human non-expert, and LLM-generated sub-corpora, the framework identifies a consistent, model-agnostic LLM epistemic signature. LLM-generated texts produce tricolon at nearly twice the expert rate ($\Delta = 0.95$), while human authors produce erotema at more than twice the LLM rate. Performed hesitancy markers appear at twice the human density in LLM output. FMD is significantly elevated in LLM texts relative to both human groups ($p < 0.001, \Delta = 0.68$), and rhetorical devices are distributed significantly more uniformly across LLM documents. The findings are consistent with theoretical intuitions derived from Gricean pragmatics, Relevance Theory, and Brandomian inferentialism. The annotation pipeline is fully automatable, making it deployable as a lightweight screening tool for epistemic miscalibration in AI-generated content and as a theoretically motivated feature set for LLM-generated text detection pipelines.
\end{abstract}


\section{Introduction}
\label{sec:intro}

Large language model (LLM) bias is a multilayered phenomenon that manifests in distinct linguistic and social dimensions \cite{gallegos_bias_2024,
ranjan_comprehensive_2024}. The existing literature predominantly addresses three critical categories: demographic and representational bias involving social hierarchies \cite{blodgett_language_2020}, factual inaccuracy often termed hallucination \cite{ji_towards_2023}, and the propagation of toxicity \cite{bender_dangers_2021}. Although these concerns are vital for model safety, they focus primarily on semantic content, i.e., the surface representation layer of LLM
responses. What remains largely unexamined are the structural metacognitive mechanisms by which models position their claims \cite{erhardt_metacognitive_2025}. Beneath explicit content lies a secondary, more subtle layer of structural bias related to the epistemic and rhetorical posture of generated text \cite{li_representations_2025}.

This deeper layer concerns the epistemic stance of a text, a concept rooted in linguistic modality that reflects a speaker's degree of
commitment to propositional content \cite{gendler_opinionated_2007,
li_linguistic_2025}. Epistemic stance is a load-bearing feature of
language that differentiates between settled knowledge, indirect
inference, and genuine uncertainty \cite{lee_are_2025,
li_representations_2025}. Rational communication relies on this
calibration to signal the reliability of evidence and the source of
information. If LLMs systematically miscalibrate these markers
relative to the actual epistemic status of their claims, they
introduce a structural bias that distorts how users assess and trust
generated content.

Recent rhetorical analyses have identified that LLMs produce
persuasive surface structures, and epistemic stance marking has
been studied as an isolated natural language processing (NLP) task \cite{clausen_hedgehunter_2010,
kramer_rhet_2025}. To the best of our knowledge, however, no
existing work has examined the relationship between rhetorical
intensity and epistemic calibration as a unified, measurable
construct \cite{erhardt_metacognitive_2025}. This gap manifests as
a decoupling in which LLMs deploy elaborate rhetorical forms, such as 
tricolon, correctio, contrastive reframing, etc, independently of
whether the claim warrants such intensity or whether epistemic
markers represent genuine uncertainty or merely performed hesitancy.
The precise divergence between rhetorical form and epistemic import
remains unmeasured.

To address this gap, we propose an epistemic-rhetorical marker
(ERM) taxonomy, a triadic framework distinguishing sentence-level
rhetorical devices, epistemic stance markers, and discourse-level
argumentative structure. The framework operationalizes metrics that quantify the decoupling
between rhetorical intensity and epistemic calibration, validated
through a corpus comprising human expert, human non-expert, and
LLM-generated writing. The approach is grounded in Gricean
pragmatics \cite{grice1975logic}, Relevance
Theory \cite{wilson2002relevance}, and Brandomian
inferentialism \cite{brandom1994making, brandom1997precis}.

This study makes three primary contributions. First, we present a theoretically grounded ERM taxonomy integrating rhetorical device analysis with epistemic modality classification across three levels of linguistic organisation: sentence-level tropes, lexical and
syntactic stance markers, and discourse-level argumentative
structure. Second, we introduce three novel corpus-applicable metrics, i.e., a form-meaning divergence (FMD) score, genuine-to-performed ratio (GPR), and rhetorical device distribution entropy (RDDE), as novel, corpus-applicable metrics for measuring epistemic-rhetorical
divergence in argumentative text. Third, we provide corpus wide empirical validation showing that LLM output exhibits systematically higher divergence and a distinct epistemic marker profile. Together, these contributions disentangle a dimension of LLM bias that has implications for AI-generated text evaluation and the computational study of epistemic stance.

\section{Related Work}
Rhetorical analysis, stylometric measurement, and epistemic marker classification have developed largely in isolation. Their structural relationship remains unmeasured. The current literature falls into three relevant domains.

\subsection{Bias in LLMs: Existing Taxonomies and Their Limits}
LLM bias research is predominantly structured around the evaluation of propositional content and surface-level
identifiers \cite{gallegos_bias_2024, ranjan_comprehensive_2024,
blodgett_language_2020}. Initial research focused heavily on
demographic and representational biases, categorising harms into allocational disparities and representational harms where specific social groups are stereotyped or
devalued \cite{raj2024breaking, raza_developing_2025}. Parallel to this, a robust body of literature addresses factual bias and hallucination \cite{ji_towards_2023, kenthapadi2024grounding}. Significant effort has also been directed toward toxicity and harmful output, developing metrics and moderation tools to identify offensive or biased language targeting marginalised communities \cite{guo_bias_2024, bender_dangers_2021,
mckee_ethics_2020}. Despite their critical importance, these
taxonomies do not analyse the structural relationship between rhetorical form and epistemic stance.

\subsection{Computational Approaches to Rhetoric and Style}
Computational stylistics establishes that individual writing style is a measurable fingerprint detectable through quantitative analysis \cite{neal2017surveying}. Stylometry leverages metrics such as Burrows' Delta to measure the distribution of most frequent words, providing a content-independent authorship signature \cite{burrows2002delta, jannidis2015improving}. This
tradition has informed recent work using lexical and syntactic features to distinguish human-written from LLM-generated text across various models \cite{agrahari_text_2025, bisztray_i_2026,
kumarage_neural_2023, zaitsu_stylometry_2025}. Computational
rhetorical analysis has sought to map discourse structure through frameworks such as rhetorical structure theory and the automated detection of complex figures and
tropes \cite{majdik_rhetoric_2024, erhardt_metacognitive_2025}.
These stylometric tools have more recently been applied to
AI-generated text detection, identifying distinctive patterns such as reduced lexical diversity and increased structural uniformity relative to human expert
writing \cite{aityan_lightweight_2026, al-shaibani_arabic_2026}.

\subsection{Epistemic Modality and Stance in NLP}
Computational work on hedging detection has provided a foundation for identifying expressions of uncertainty, particularly within scientific literature \cite{clausen_hedgehunter_2010,
medlock_weakly_2007}. Hedge cues have been shown to be
high-precision markers of uncertainty, though their detection remains highly domain-dependent \cite{szarvas_hedge_2008, li_linguistic_2025}. In the context of LLMs, epistemic stance classification has evolved into the study of honesty alignment, where models are trained to verbalise confidence levels and express
uncertainty explicitly~\cite{clark_epistemic_2025, lee_are_2025, tao_can_2025}. No existing work correlates the presence of these markers with the rhetorical elaborateness of the surrounding text. Moreover, performed hesitancy is not distinguished from genuine
epistemic marking as a functional category.

\section{Theoretical Framework}
\label{sec:theoretical_framework}

\subsection{Foundational Premises}
\label{sec:foundations}

We hypothesize that the decoupling of rhetorical form from epistemic grounding is a measurable property of text. Three theoretical traditions, i.e., Gricean pragmatics, Relevance Theory, and Brandomian inferentialism, provide a unified account of why this relationship is normatively governed and why its disruption leaves a recoverable structural signal.

\subsubsection{Gricean Cooperative Principle}
\label{sec:grice}

Gricean pragmatics holds that speakers in a communicative exchange are expected to observe implicit maxims governing the quantity, quality, relation, and manner of their
contributions~\cite{grice1975logic}. The maxims of quantity and quality together establish a normative expectation that the \textit{weight} of an assertion, the degree of confidence and elaborateness with which it is presented, should be proportionate to the speaker's actual epistemic position. A rhetorically elaborate claim that is evidentially thin violates the normative structure that
makes assertion a cooperative act. Consider the following pair:

\begin{quote}
(1)\; \textit{``Some studies suggest that sleep deprivation may impair consolidation of declarative memory.''}\\[4pt]
(2)\; \textit{``Without question, sleep deprivation impairs
attention, corrupts consolidation, and destroys the brain's capacity to retain what it has learned across every domain of cognitive function, with consequences that are nothing short of profound.''}
\end{quote}

\noindent Claim (1) deploys modal hedging (\textit{may}), an
evidential restrictor (\textit{some studies suggest}), and domain qualification (\textit{declarative memory}); its rhetorical weight is proportionate to the epistemic position it reports. Claim (2) deploys tricolon, categorical assertion (\textit{without question}),
and auxesis (\textit{nothing short of profound}). Its rhetorical weight vastly exceeds any single body of evidence that could license it. A reader encountering Claim (2) is misled not by the content but by the assertoric form in which it is packaged.

\subsubsection{Relevance Theory}
\label{sec:relevance-theory}

Relevance Theory grounds the Gricean intuition in a cognitive account of comprehension \cite{wilson2002relevance}. It proposes
that communicative behaviour is governed by a presumption of optimal relevance: when a speaker produces an utterance requiring interpretive effort, that effort creates a corresponding expectation of proportionate cognitive effects. Rhetorical devices such as tricolon or auxesis impose structural complexity on the reader. Relevance Theory predicts this cost is licensed only when the
content warrants it. Consider a passage from an LLM-generated example policy report n response to a prompt:

\begin{quote}
(3)\; \textit{``Across the full spectrum of deployment contexts, from hiring to healthcare, from credit scoring to criminal justice, and from content moderation to educational assessment, algorithmic systems raise profound, multi-dimensional, and deeply interconnected questions about equity.''}
\end{quote}

\noindent A reader is expected to track a six-item enumeration, resolve three parallel prepositional phrases, and integrate the import of \textit{profound}, \textit{multi-dimensional}, and \textit{deeply interconnected}. What the sentence delivers is the proposition that algorithmic systems raise questions about equity, a claim any
reader already accepts. The cognitive yield is arguably nearly zero relative to the processing cost, and this gap is computationally measurable.

\subsubsection{Brandomian Inferentialism}
\label{sec:inferentialism}

Brandom's account of discursive practice treats a speaker's
assertion as the undertaking of a social and normative
commitment \cite{brandom1994making, brandom1997precis}. The
\textit{strength} of that commitment is a normatively governed function of the \textit{entitlements} the speaker actually possesses. When assertoric strength exceeds the entitlements backing it, as in Claim (2), a reader is licensed to treat the overclaimed proposition as a warranted premise for subsequent reasoning, concluding that the effect size is large, the evidence settled, and the claim universal. The speaker has propagated an epistemic deficit into inference chains they will never be held
accountable for. It is this downstream liability that gives the form-meaning decoupling its normative weight and motivates the divergence score as an instrument for detecting structural bias rather than merely poor style.

\subsection{The ERM Taxonomy}
\label{sec:erm}

\begin{figure}[htbp]
  \centering
  \includegraphics[width=\textwidth]{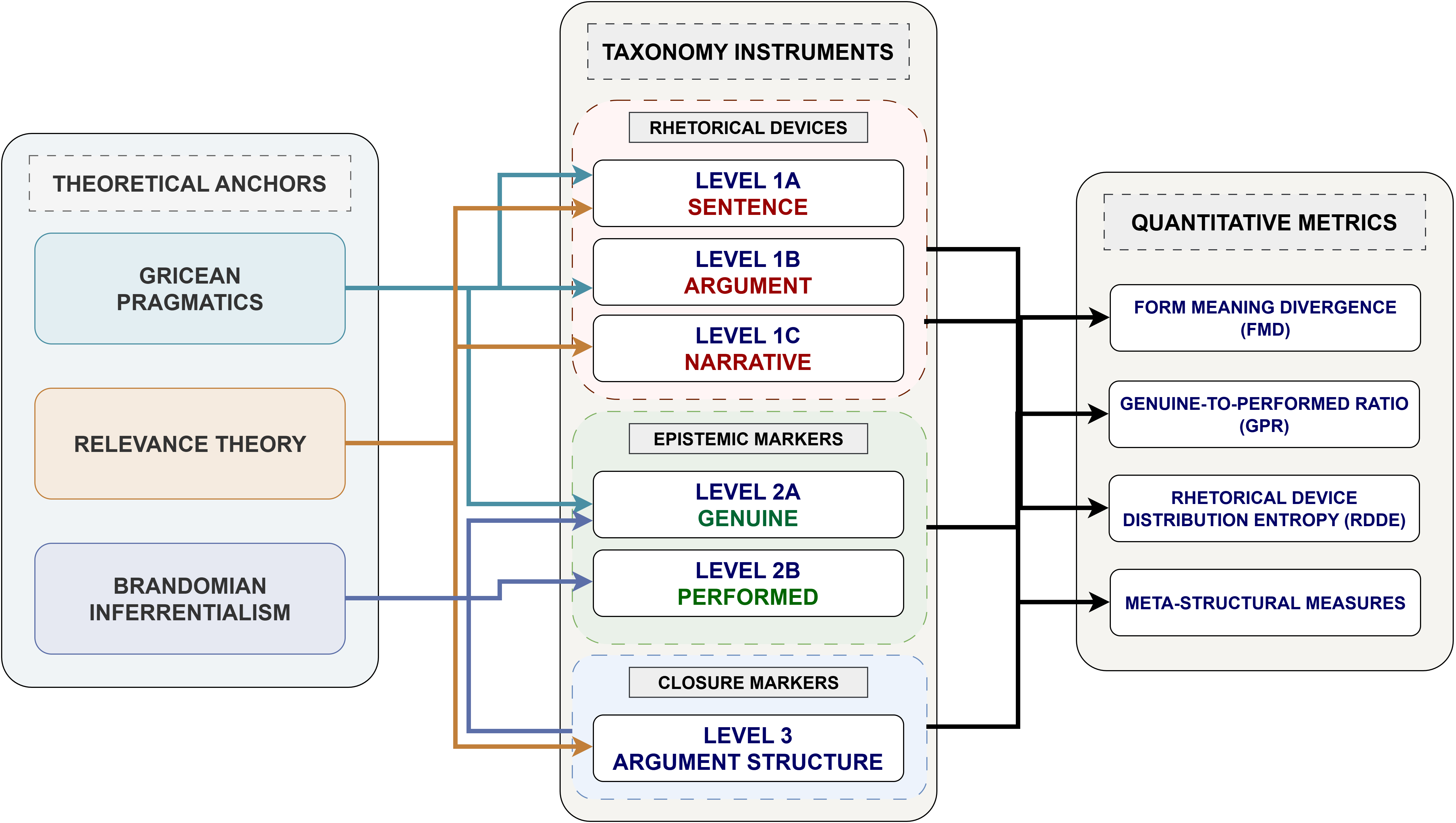}
  \caption{Architecture of the ERM taxonomy showing linkages
  between the three theoretical anchors (left), the six taxonomy levels (centre), and the four types of quantitative metrics (right).}
  \label{fig:erm_architecture}
\end{figure}

The ERM taxonomy is proposed to operationalize rhetorical form and epistemic grounding as independently annotatable properties of text across three separable levels of linguistic organization. Its architecture, theoretical anchors, and derived metrics are illustrated in Figure \ref{fig:erm_architecture}. The complete inventory is summarized in Table \ref{tab:erm_summary}, with full definitions and annotated examples in \ref{app:erm}.

\begin{table}[htbp]
\centering
\caption{Overview of the ERM taxonomy; full definitions and
annotated examples are in \ref{app:erm}.}
\label{tab:erm_summary}
\vspace{4pt}
\small
\setlength{\tabcolsep}{8pt}
\renewcommand{\arraystretch}{1.4}
\begin{tabularx}{\textwidth}{@{}
  >{\RaggedRight\arraybackslash}p{2.2cm}
  >{\RaggedRight\arraybackslash}p{3.2cm}
  >{\RaggedRight\arraybackslash}X@{}}
\toprule
\textbf{Level} & \textbf{Sub-level} & \textbf{Devices / Markers} \\
\midrule
\multirow{3}{2.8cm}{\textbf{Level 1}\\[2pt]Rhetorical Devices}
  & 1a:\enspace Sentence
  & Tricolon, Anaphora, Chiasmus, Erotema, Sententia \\
  \cmidrule(l){2-3}
  & 1b:\enspace Argument
  & Correctio, Enumeratio, Auxesis \\
  \cmidrule(l){2-3}
  & 1c:\enspace Narrative
  & Peripeteia, Expectation-Violation-Resolution \\
\midrule
\multirow{2}{2.8cm}{\textbf{Level 2}\\[2pt]Epistemic Stance Markers}
  & 2a:\enspace Genuine
  & Modal Auxiliaries, Adverbial Expressions,
    Syntactic Restrictor Phrases, Evidential Markers \\
  \cmidrule(l){2-3}
  & 2b:\enspace Performed
  & Complexity Tokens, Meta-epistemic Gestures \\
\midrule
\multirow{1}{2.8cm}{\textbf{Level 3}\\[2pt]Argumentative Structure}
  & \multicolumn{1}{l}{}
  & Aporetic Endpoint, Synthetic Closure,
    Premature Closure, Speculative Depth
\end{tabularx}
\end{table}

\subsubsection{Level 1 -- Rhetorical Devices}
\label{sec:level1}
Level 1 captures formal presentational structures that modulate expressive intensity, i.e., the \textit{how} rather than the \textit{what} of a claim. Grounded in the Gricean quantity maxim and Relevance Theory's processing-cost principle, it provides the rhetorical intensity component for computing FMD and RDDE. Ten devices are organized across three scales of operation
(Tables \ref{tab:erm_1a}--\ref{tab:erm_1c}).

\subsubsection{Level 2 -- Epistemic Stance Markers}
\label{sec:level2}
Level 2 classifies lexical and syntactic devices by which a text encodes its assertoric commitment. The central distinction is between \textit{genuine epistemic markers} (2a) that ground a claim in an identifiable evidential base \cite{gendler_opinionated_2007} and \textit{performed hesitancy} (2b) that adopt the surface register of uncertainty without its formal apparatus. Both levels
feed FMD and GPR differently (see Section \ref{sec:methodology}), an opposition that operationalizes the Brandomian distinction between inferential entitlement and its simulation \cite{brandom1997precis}.

\subsubsection{Level 3 -- Discourse-level Argumentative Structure}
\label{sec:level3}
Level~3 characterizes the global argumentative endpoint and
inferential trajectory of a text, distinguishing four markers as defined in Table \ref{tab:erm_3}. These are annotated at document level and reported as sub-corpus proportions.

\section{Methodological Implementation}
\label{sec:methodology}

\begin{figure}[htbp]
\centering
\includegraphics[width=\textwidth]{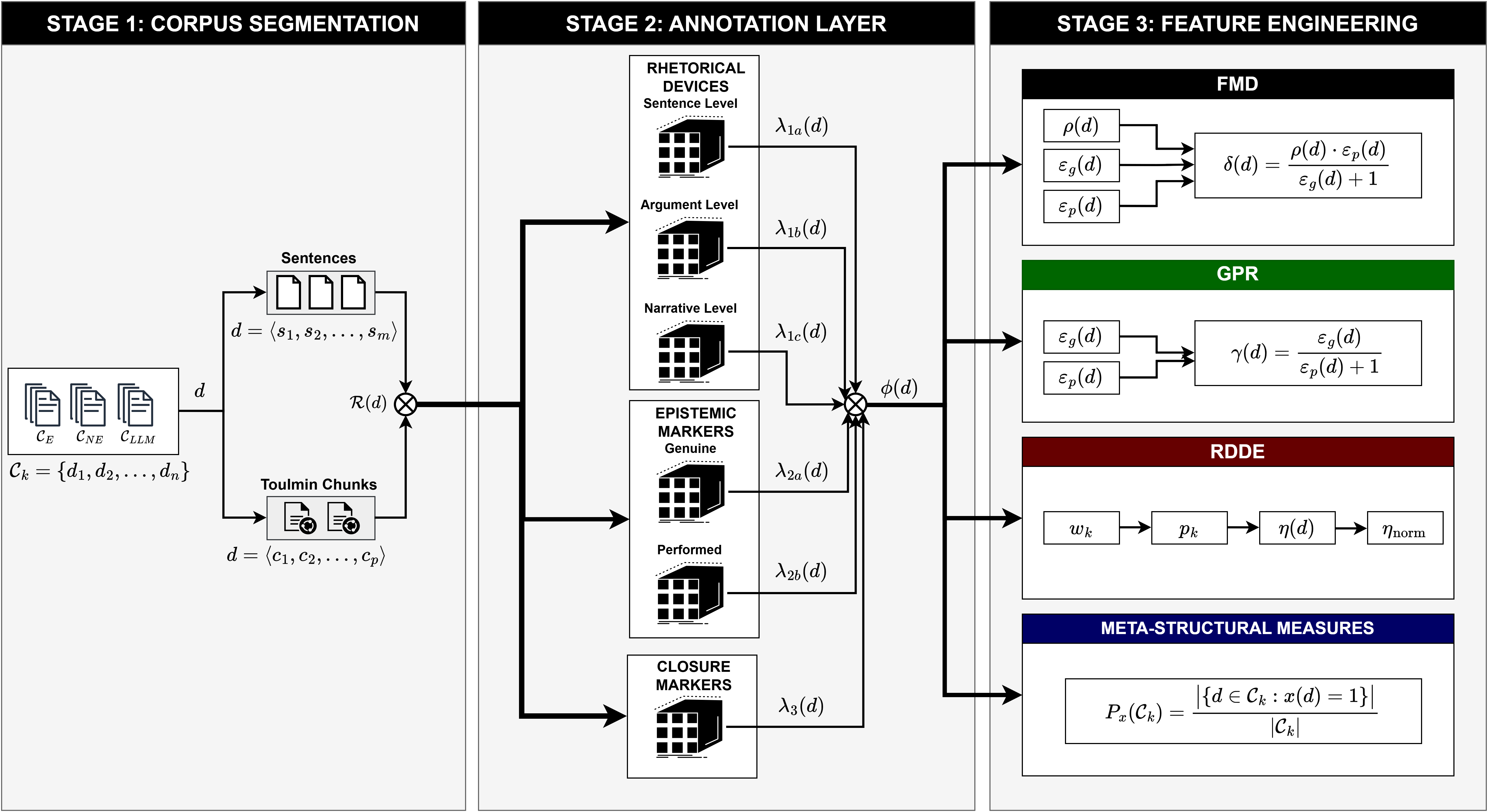}
\caption{The ERM architecture pipeline. Stage~1 constructs the corpus and segments each document into sentence sequences and Toulmin semantic chunks. Stage~2 applies the ERM taxonomy across five parallel annotation passes covering rhetorical devices at three scales, genuine and performed epistemic markers, and closure markers. Stage~3 transforms the annotation output into three composite metrics, i.e., FMD, GPR, RDDE and four discourse-level structural proportions.}
\label{fig:pipeline}
\end{figure}

The proposed methodological pipeline, illustrated in
Figure~\ref{fig:pipeline}, comprises three stages: corpus construction and segmentation, annotation, and ERM feature engineering.\footnote{All annotation prompts, corpus metadata, and analysis scripts will be open-sourced (post-publication) to support reproducibility and replication across extended and more diverse corpora.}

\subsection{Stage 1: Corpus Construction and Segmentation}
\label{sec:stage1}
The corpus comprises three sub-corpora corresponding to the three author types under investigation. Let the full corpus be defined as

\begin{equation}
\mathcal{C} = \{\mathcal{C}_E,\; \mathcal{C}_{NE},\; \mathcal{C}_{LLM}\}
\label{eq:corpus}
\end{equation}
\noindent where $\mathcal{C}_{E}$, $\mathcal{C}_{NE}$, and
$\mathcal{C}_{LLM}$ denote the human expert, human non-expert, and LLM-generated sub-corpora respectively. Each sub-corpus contains $|\mathcal{C}_k| = 75$ documents:
\begin{equation}
\mathcal{C}_k = \{d_1, d_2, \ldots, d_{75}\}, \quad
k \in \{E, NE, LLM\}
\label{eq:subcorpus}
\end{equation}

yielding a total corpus of 225 documents. The sub-corpora are summarized in Table~\ref{tab:corpus_summary}.

\begin{table}[htbp]
\centering
\caption{Corpus summary ($n = 75$ per sub-corpus, 225 total).
A pre-November 2022 cutoff is applied to \textsc{he} and
\textsc{hn} to exclude texts potentially assisted by generative
AI tools. Each \textsc{lg} text was generated using a uniform
prompt template comprising an open-ended argumentative question
derived from the corresponding \textsc{he} text's central topic,
a 1500-word length target, and fixed structural instructions to
take a clear position, acknowledge a counterargument, and engage
honestly with uncertainty.}
\label{tab:corpus_summary}
\vspace{2pt}
\small
\setlength{\tabcolsep}{5pt}
\renewcommand{\arraystretch}{1.3}
\begin{tabularx}{\textwidth}{@{}
  >{\RaggedRight\arraybackslash}p{2.4cm}
  >{\RaggedRight\arraybackslash}p{2.0cm}
  >{\centering\arraybackslash}p{1.4cm}
  >{\RaggedRight\arraybackslash}X
  >{\RaggedRight\arraybackslash}X@{}}
\toprule
\textbf{Sub-corpus} &
\textbf{Period} &
\textbf{Mean sent.} &
\textbf{Sources} &
\textbf{Topical scope} \\
\midrule
Human Expert (HE)
  & 2011--2022
  & 92.8
  & Aeon Magazine; CATO Institute; SSIR; ResearchGate;
    \textit{The New Yorker}; \textit{The Guardian};
    Nautilus; Wired; Psyche
  & Philosophy of mind; philosophy and ethics; psychology;
    society and culture; technology policy;
    organisational communication \\
Human Non-Expert (HN)
  & 2016--2022
  & 96.7
  & Real Life Magazine; Medium
  & Same topical scope as HE; non-academic,
    non-institutional authors \\
LLM-Generated (LG)
  & 2024--2025
  & 104.7
  & GPT-5.1; DeepSeek-V3.2; Claude-Sonnet-4.6;
    Gemini-2.5-Pro
  & Prompts derived from corresponding HE texts \\
\bottomrule
\end{tabularx}
\end{table}

\subsubsection{Segmentation}
\label{sec:segmentation}

Each document $d$ is segmented into two distinct representational units serving different levels of ERM annotation, applied uniformly across all three sub-corpora.

For Levels~1 and~2, each document $d$ is represented as an ordered sequence of sentences produced by spaCy sentence boundary detection \cite{honnibal2020spacy}:

\begin{equation}
d = \langle s_1, s_2, \ldots, s_m \rangle
\label{eq:sentence_seq}
\end{equation}
Rhetorical devices and epistemic stance markers are annotated at this level.

For Level~3, each document $d$ is represented as an ordered sequence of argumentative chunks:

\begin{equation}
d = \langle c_1, c_2, \ldots, c_p \rangle
\label{eq:chunk_seq}
\end{equation}

where each chunk $c_k$ is a contiguous span of one or more sentences assigned a single argumentative function label drawn from a seven-way typology adapted from Toulmin's model of argumentation \cite{toulmin2003uses}:

\begin{multline}
\tau(c_k) \in \{\textsc{Claim},\, \textsc{Grounds},\, \textsc{Warrant},\,
\textsc{Backing},\, \\
\textsc{Qualifier},\, \textsc{Rebuttal},\, \textsc{Non-Argumentative}\}
\label{eq:toulmin_labels}
\end{multline}

\textsc{Non-Argumentative} is assigned to transitional, expository, definitional, or illustrative spans that carry no direct argumentative function, following the precedent established in computational argumentation mining \cite{stab2017parsing}. Semantic chunks are non-overlapping and jointly exhaustive over the document:

\begin{equation}
\bigcup_{k=1}^{p} c_k = d, \qquad c_k \cap c_{k'} = \emptyset
\;\; \text{for} \;\; k \neq k'
\label{eq:chunk_partition}
\end{equation}

Chunk boundaries are identified using a locally hosted LLM to segment each document and return structured span indices with type labels. \textsc{Non-Argumentative} chunks are excluded from Level~3 annotation; the remaining chunks constitute the argumentative skeleton of the document used to contextualise endpoint and trajectory judgements.

The complete segmentation yields two parallel representations for each document $d \in \mathcal{C}$:
\begin{equation}
\mathcal{R}(d) = \bigl(\langle s_1, \ldots, s_m \rangle,\;
\langle c_1, \ldots, c_p \rangle\bigr)
\label{eq:representation}
\end{equation}

\subsection{Stage 2: Annotation}
\label{sec:stage2}

For each document $d$, the annotator processes each text unit in a separate pass and returns a binary judgement for each marker in the corresponding taxonomy level, producing the structured annotation vector:

\begin{equation}
\phi(d) = \bigl(\lambda_{1a}(d),\;\lambda_{1b}(d),\;\lambda_{1c}(d),\;
\lambda_{2a}(d),\;\lambda_{2b}(d),\;\lambda_{3}(d)\bigr)
\label{eq:annotation_vector}
\end{equation}

\noindent where $\lambda_{1a}(d)$, $\lambda_{2a}(d)$, and
$\lambda_{2b}(d)$ are binary matrices of shape $m_d \times 5$, $m_d \times 4$, and $m_d \times 2$ respectively, recording sentence-level rhetorical devices and epistemic stance markers; $\lambda_{1b}(d)$ is a binary matrix of shape $p \times 3$ recording argument-level rhetorical devices per chunk; and $\lambda_{1c}(d)$ and $\lambda_{3}(d)$ are binary vectors of shape $1 \times 2$ and $1 \times 4$ recording narrative-level rhetorical devices and Level~3 argumentative structure markers at document level, with the Toulmin chunk structure from Stage~1 providing structured context for Level~3 judgements.

Algorithm~\ref{alg:erma} summarises the complete ERM pipeline. Each stage is designed to be executed either by a trained human annotator or by an LLM. Human annotation is feasible, albeit requiring training on the ERM taxonomy, familiarity with Toulmin argumentation paradigm, and close reading (Level 3). LLM annotation may employ a capable, decent sized model. The choice between human and LLM annotation trades interpretive depth for throughput while both modalities ultimately producing the annotation vector $\phi(d)$ required
by Stage~3.

\begin{algorithm}[tbp]
\caption{Automated ERM Pipeline}
\label{alg:erma}
\begin{algorithmic}[1]

\STATE \textbf{Input:} Corpus $\mathcal{C} = \{\mathcal{C}_E, \mathcal{C}_{NE}, \mathcal{C}_{LLM}\}$
\STATE \textbf{Output:} Per-document features $\{\delta(d), \gamma(d), \eta_{\text{norm}}(d)\}$; sub-corpus proportions $P_x(\mathcal{C}_k)$

\vspace{0.5em}
\STATE \textbf{Stage 1: Segmentation}
\FORALL{document $d \in \mathcal{C}$}
    \STATE Detect sentence boundaries $\langle s_1,\ldots,s_m \rangle$ via spaCy
    \STATE Segment $d$ into Toulmin chunks $\langle c_1,\ldots,c_p \rangle$ with labels $\tau(c_k)$
\ENDFOR

\vspace{0.5em}
\STATE \textbf{Stage 2: Annotation} (human or LLM annotator)
\FORALL{document $d \in \mathcal{C}$}
    \STATE \textbf{Pass 1:} annotate each $s_i \rightarrow \lambda_{1a}(d)$
    \STATE \textbf{Pass 2:} annotate each $s_i \rightarrow \lambda_{2a}(d), \lambda_{2b}(d)$
    \STATE \textbf{Pass 3:} annotate each non-\textsc{Non-Arg} $c_k \rightarrow \lambda_{1b}(d)$
    \STATE \textbf{Pass 4:} annotate $d \rightarrow \lambda_{1c}(d), \lambda_3(d)$
\ENDFOR

\vspace{0.5em}
\STATE \textbf{Stage 3: Feature Engineering}
\FORALL{document $d \in \mathcal{C}$}
    \STATE Compute $\rho(d)$, $\varepsilon_g(d)$, $\varepsilon_p(d)$ from $\lambda_{1*}(d)$, $\lambda_{2*}(d)$, and $m_d$
    \STATE Compute $\delta(d) \leftarrow (11)$
    \STATE Compute $\gamma(d) \leftarrow (12)$
    \STATE Partition $d$ into 50-word windows
    \STATE Compute $\eta_{\text{norm}}(d) \leftarrow (14)$
    \STATE Extract binary Level 3 indicators $\sigma(d), \alpha(d), \pi(d), \varsigma(d)$
\ENDFOR

\FORALL{sub-corpus $\mathcal{C}_k$}
    \STATE Compute $P_x(\mathcal{C}_k)$ for each $x \in \{\sigma, \alpha, \pi, \varsigma\}$
\ENDFOR

\end{algorithmic}
\end{algorithm}

\subsection{Stage 3: ERM Feature Engineering}
\label{sec:stage3}
The annotation matrices in $\phi(d)$ are transformed into four
features, each capturing a distinct dimension of
epistemic-rhetorical miscalibration. All density measures are
normalized per sentence given $m_d$, the sentence count of
document $d$.
\subsubsection{Form-Meaning Divergence (FMD)}
\label{sec:fmd}
In order to measure the degree to which rhetorical elaboration exceeds epistemic grounding, the total rhetorical device count is aggregated across all three Level~1 sub-levels and normalized as
\begin{equation}
\rho(d) = \frac{\sum\lambda_{1a}(d) + \sum\lambda_{1b}(d) +
\sum\lambda_{1c}(d)}{m_d}
\label{eq:rho}
\end{equation}
The densities of genuine and performed epistemic markers are defined analogously as
\begin{equation}
\varepsilon_g(d) = \frac{\sum\lambda_{2a}(d)}{m_d},
\qquad
\varepsilon_p(d) = \frac{\sum\lambda_{2b}(d)}{m_d}
\label{eq:eps}
\end{equation}
Form-Meaning Divergence is then computed as
\begin{equation}
\delta(d) = \frac{\rho(d) \cdot \varepsilon_p(d)}{\varepsilon_g(d) + 1}
\label{eq:fmd}
\end{equation}
\noindent The multiplicative structure ensures that divergence is
only elevated when rhetorical elaboration and epistemic
miscalibration co-occur. $\varepsilon_g(d)$ in the denominator
attenuates the score proportionally to the genuine epistemic grounding. The constant $+1$ prevents the division by zero.
\subsubsection{Genuine-to-Performed Epistemic Ratio (GPR)}
\label{sec:gpr}
A text may exhibit low divergence simply because its rhetorical
intensity is low, while still relying predominantly on performed rather than genuine epistemic marking. Therefore, epistemic calibration can be isolated independently of rhetorical intensity as
\begin{equation}
\gamma(d) = \frac{\varepsilon_g(d)}{\varepsilon_p(d) + 1}
\label{eq:gpr}
\end{equation}
\noindent where $\gamma(d) > 1$ indicates that genuine markers outweigh performed markers and $\gamma(d) < 1$ indicates the reverse. Whereas $\delta(d)$ captures the relationship between rhetorical form and epistemic content, $\gamma(d)$ captures the internal composition of the epistemic layer alone.
\subsubsection{Rhetorical Device Distribution Entropy (RDDE)}
\label{sec:rdde}
It is reasonable to assume that human expert writers cluster rhetorical devices around moments of genuine argumentative stress. Hence, a uniform distribution across
the text suggests deployment driven by local stylistic habit rather than argumentative logic. The document is partitioned into windows of up to 50 words; letting $w_k$ denote the total Level~1 device count in window $k$, the Shannon entropy over the device distribution is:
\begin{equation}
\eta(d) = -\sum_{k=1}^{K} p_k \log_2 p_k, \qquad
p_k = \frac{w_k}{\sum_{k'=1}^{K} w_{k'}}
\label{eq:rdde}
\end{equation}
To ensure comparability across documents of different lengths, $\eta(d)$ is normalised by the maximum possible entropy for $K$
windows:
\begin{equation}
\eta_{\text{norm}}(d) = \frac{\eta(d)}{\log_2 K}
\label{eq:rdde_norm}
\end{equation}
\noindent where $\eta_{\text{norm}}(d) = 1$ indicates a perfectly uniform device distribution and $\eta_{\text{norm}}(d) \to 0$ indicates maximal clustering around argumentative pressure points.
\subsubsection{Discourse-Level Meta-Structural Measures}
\label{sec:level3_features}
Level~3 annotation yields four binary document-level indicators drawn from the ERM taxonomy: synthetic closure $\sigma(d)$, aporetic endpoint $\alpha(d)$, premature closure $\pi(d)$, and speculative depth $\varsigma(d)$, each $\in \{0,1\}$. These are corpus-level proportions and are comparable across sub-corpora using chi-squared test. For each
marker $x \in \{\sigma, \alpha, \pi, \varsigma\}$, the
sub-corpus proportion is computed as
\begin{equation}
P_x(\mathcal{C}_k)=\frac{\bigl|\{d \in \mathcal{C}_k:x(d)=1\}\bigr|}{|\mathcal{C}_k|}
\label{eq:level3_prop}
\end{equation}

\section{Results and Discussion}
\label{sec:results}

Table~\ref{tab:overview} presents a broad statistical overview of
the ERM metrics and key device markers across the three sub-corpora. Several patterns are immediately apparent.
LLM-generated texts show the highest FMD
($\bar{\delta} = 0.017$) and the most uniform RDDE ($\bar{\eta}_{\text{norm}} = 0.753$), while human expert texts lead on the GPR ($\bar{\gamma} = 0.267$). At the device level, tricolon is the strongest single differentiator, with LLM-generated texts
producing nearly twice the expert rate. The erotema, interestingly, shows the reverse pattern, being substantially higher in both human groups than in LLM output. Performed epistemic marker density is approximately double in
LLM texts relative to either human group, while the two human sub-corpora are statistically indistinguishable on this measure. Within the LLM sub-corpus, no significant differences are observed across the four models tested, suggesting that the miscalibration pattern is a structural property of LLM
generation rather than a model-specific artifact.

\begin{table}[htbp]
\centering
\caption{Key ERM metrics and device markers across sub-corpora. Bold values mark sub-corpora that differ significantly from at least one other group (significance threshold $p < 0.05$, corrected for multiple
comparisons). $\Delta$ is Cohen's effect size for the Human Expert vs LLM-Generated comparison.}
\label{tab:overview}
\vspace{4pt}
\small
\setlength{\tabcolsep}{5pt}
\renewcommand{\arraystretch}{1.35}
\begin{tabularx}{\textwidth}{@{}
  >{\RaggedRight\arraybackslash}p{3.4cm}
  >{\centering\arraybackslash}X
  >{\centering\arraybackslash}X
  >{\centering\arraybackslash}X
  >{\centering\arraybackslash}p{1.0cm}@{}}
\toprule
\textbf{Metric} &
\textbf{HE} $\mu\,(\sigma)$ &
\textbf{HN} $\mu\,(\sigma)$ &
\textbf{LG} $\mu\,(\sigma)$ &
$\Delta$ \\
\midrule
\multicolumn{5}{@{}l}{\textit{Composite metrics}} \\[2pt]
FMD $\delta(d)$
  & 0.009 (0.010) & 0.012 (0.019) & \textbf{0.017 (0.016)} & 0.68 \\
GPR $\gamma(d)$
  & \textbf{0.267 (0.133)} & 0.172 (0.075) & 0.217 (0.105) & 0.42 \\
RDDE $\eta_{\text{norm}}(d)$
  & 0.666 (0.143) & 0.697 (0.123) & \textbf{0.753 (0.083)} & 0.74 \\
\midrule
\multicolumn{5}{@{}l}{\textit{Level~1 device markers (mean count per document)}} \\[2pt]
Tricolon
  & 3.73 (3.48) & 4.87 (3.26) & \textbf{7.13 (3.66)} & 0.95 \\
Erotema
  & \textbf{5.55 (5.99)} & \textbf{5.11 (5.03)} & 2.28 (2.17) & 0.73 \\
Correctio
  & 0.40 (0.64) & 0.45 (0.83) & 0.17 (0.48) & 0.40 \\
\midrule
\multicolumn{5}{@{}l}{\textit{Level~2 epistemic markers}} \\[2pt]
Performed $\varepsilon_p$
  & 0.057 (0.051) & 0.058 (0.069) & \textbf{0.114 (0.102)} & 0.72 \\
Complexity tokens
  & 4.63 (4.96) & 4.77 (7.03) & \textbf{7.33 (7.33)} & 0.43 \\
\bottomrule
\end{tabularx}
\end{table}

\subsection{Rhetorical Device Repertoire}
\label{sec:results_devices}

\begin{figure}[htbp]
\centering
\includegraphics[width=\textwidth]{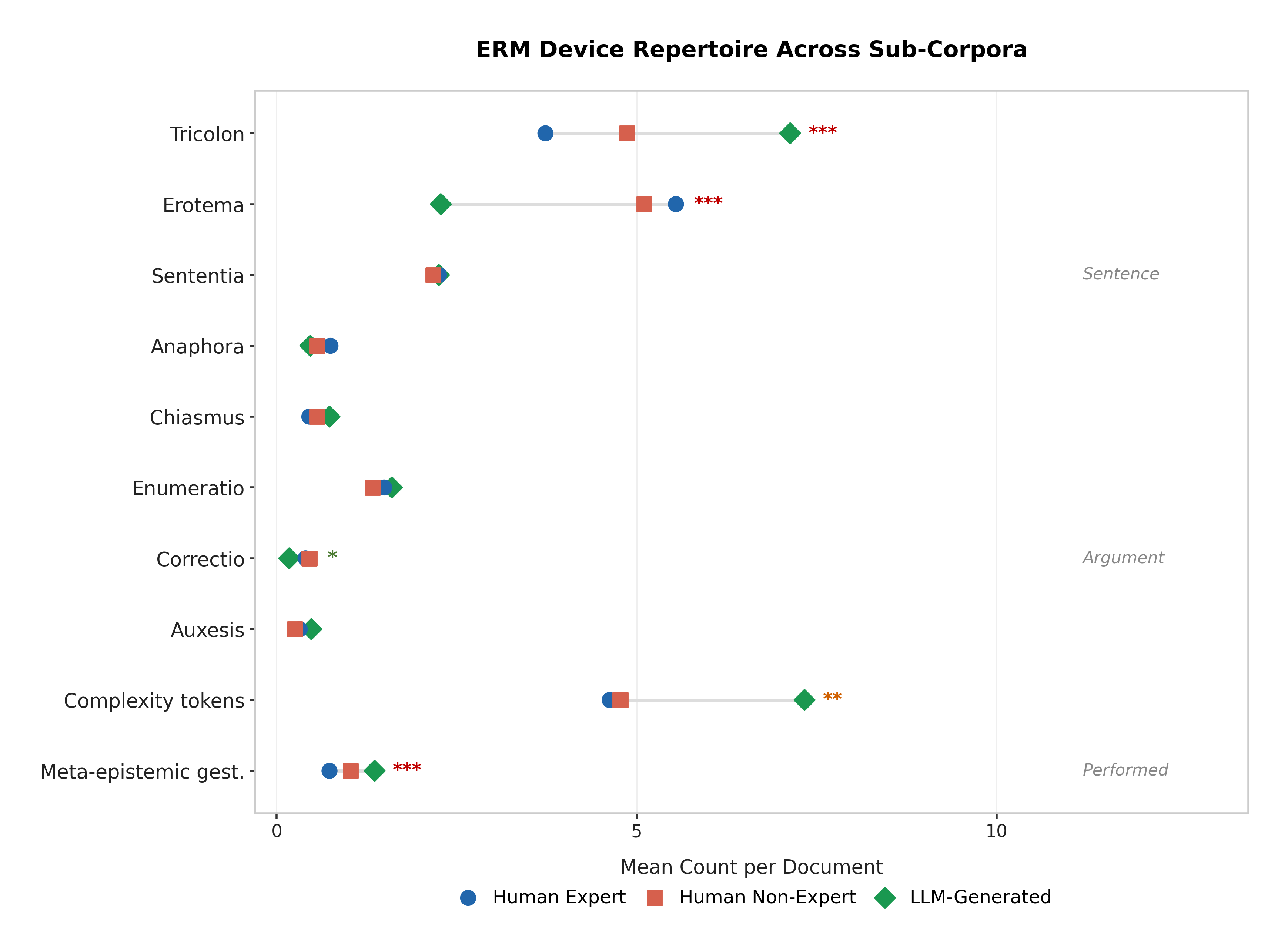}
\caption{Mean count per document of all Level~1 and Level~2b markers. Significance markers: $^{*}p{<}.05$, $^{**}p{<}.01$, $^{***}p{<}.001$. Unmarked devices show no significant difference. Section labels on the right indicate the ERM taxonomy level.}
\label{fig:device_repertoire}
\end{figure}

Figure~\ref{fig:device_repertoire} reveals a systematic divergence
in rhetorical devices between LG and human-authored texts. \textbf{Tricolon} is the strongest differentiating marker ($p < 0.001$) with LLM-generated texts producing a mean of 7.13 per document versus 3.73 for human experts ($\Delta$ $= 0.95$), thus consistent with tricolon
functioning as a default structural filler deployed independently
of argumentative occasion. 

\textbf{Erotema} shows the opposite
pattern. LLM texts show a frequency of only 2.28 per document against 5.55 (HE) and 5.11 (HNE). While both human groups are  statistically indistinguishable from each other ($p = 1.00$), these are significantly higher than LG ($\Delta \approx 0.73$). Since erotema requires the assertion of a proposition through the form of a question, its suppression in LLM output is consistent with occasion-dependent devices being less accessible to statistical generation, though prompt-induced register effects cannot be fully excluded. 

Both \textbf{performed epistemic markers} are significantly elevated in LLM output relative to both human groups. The observation confirms performed hesitancy as a specific LLM signature. \textbf{Correctio} is modestly
but significantly lower in LLM texts ( $\Delta \approx 0.41$),
consistent with reduced genuine argumentative revision. Other markers show no significant differences
across sub-corpora.

\subsection{Epistemic Positioning}
\label{sec:results_epistemic}
\begin{figure}[htbp]
\centering
\includegraphics[width=\textwidth]{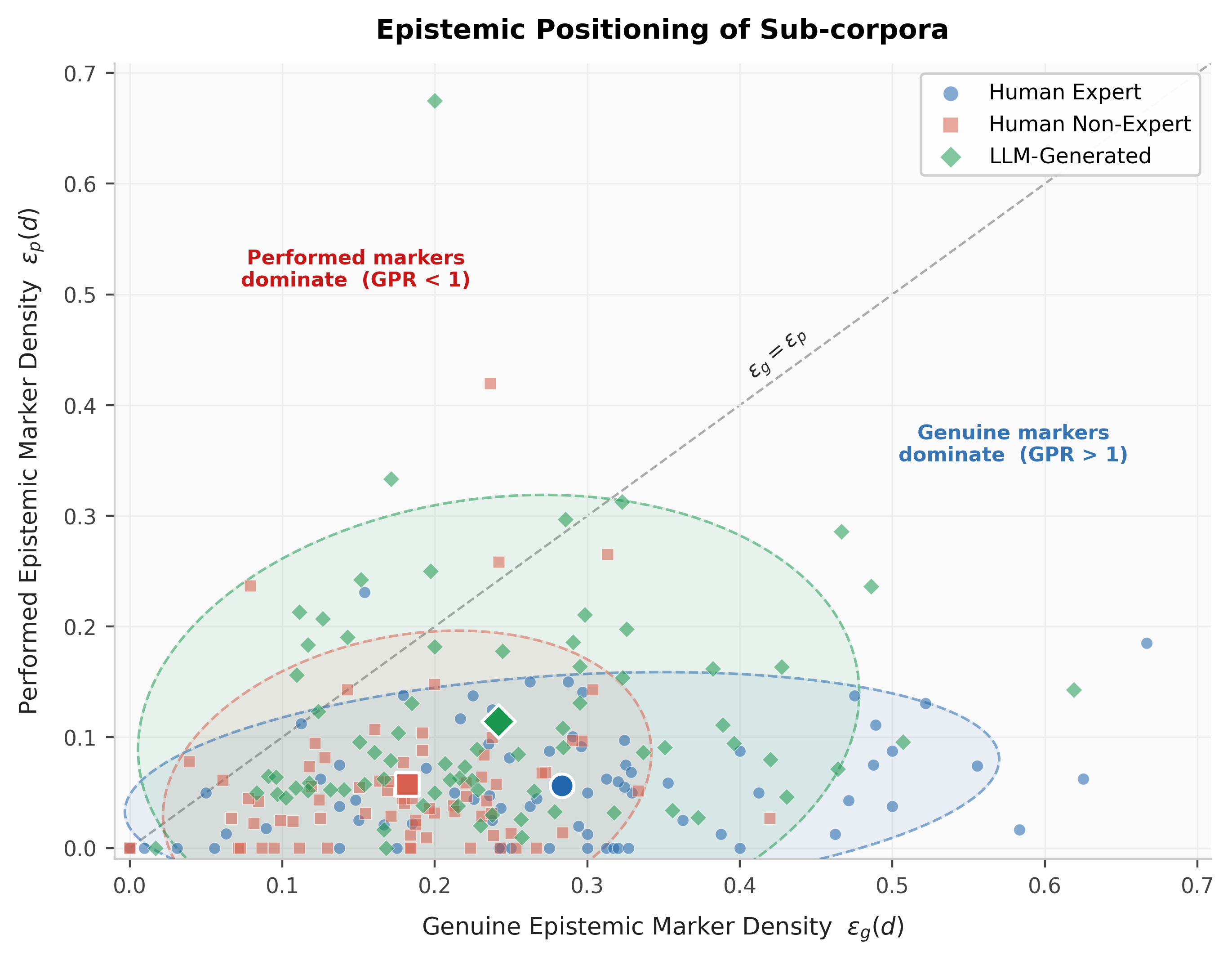}
\caption{Distribution of documents in the genuine ($\varepsilon_g$)
versus performed ($\varepsilon_p$) epistemic marker density plane.
Large markers denote group centroids. Dashed ellipses are 95\%
confidence regions. The diagonal reference line marks
$\varepsilon_g = \varepsilon_p$ (GPR~$= 1$); documents above the
line show performed-marker dominance and documents below show
genuine-marker dominance.}
\label{fig:epistemic_scatter}
\end{figure}

The mapping of each document in the plane $\varepsilon_g$ -- $\varepsilon_p$ reveals three distinct epistemic positions (Figure~\ref{fig:epistemic_scatter}). The HE ellipse extends farthest along the $\varepsilon_g$ axis, with the centroid located well into the genuine-dominant region. The compressed and origin-positioned HN ellipse reflects low density of both marker types. The LG  ellipse is notably elongated along the $\varepsilon_p$ axis, with the centroid
displaced upward relative to both human groups, indicating systematically elevated performed marker density relative to genuine epistemic grounding.

The performed epistemic marker density of LG is significantly higher  ($\bar{\varepsilon}_p = 0.114$) as compared to HE ($\bar{\varepsilon}_p = 0.057$; $\Delta = 0.72$, $p < 0.001$) and HN ($\bar{\varepsilon}_p = 0.058$; $\Delta = 0.65$, $p < 0.001$) texts. The near-identical $\varepsilon_p$ values of the two human groups confirm that performed hesitancy is a specifically LLM signature rather than a marker of non-expertise.

The density of genuine epistemic markers $\varepsilon_g$ is highest in HE texts ($\bar{\varepsilon}_g = 0.283$), followed by LG ($\bar{\varepsilon}_g = 0.242$) and HN texts ($\bar{\varepsilon}_g = 0.182$). This ordering of LLM lying between the two human groups seems consistent with the corpus design, i.e., the uniform prompt instruction to \textit{engage honestly with uncertainty} likely induced more genuine epistemic marking in LLM output than unconstrained generation might have produce.

\subsection{Composite Metrics}
\label{sec:results_composite}

\subsubsection{Form-Meaning Divergence}
\label{sec:results_fmd}

\begin{figure}[htbp]
\centering
\includegraphics[width=\textwidth]{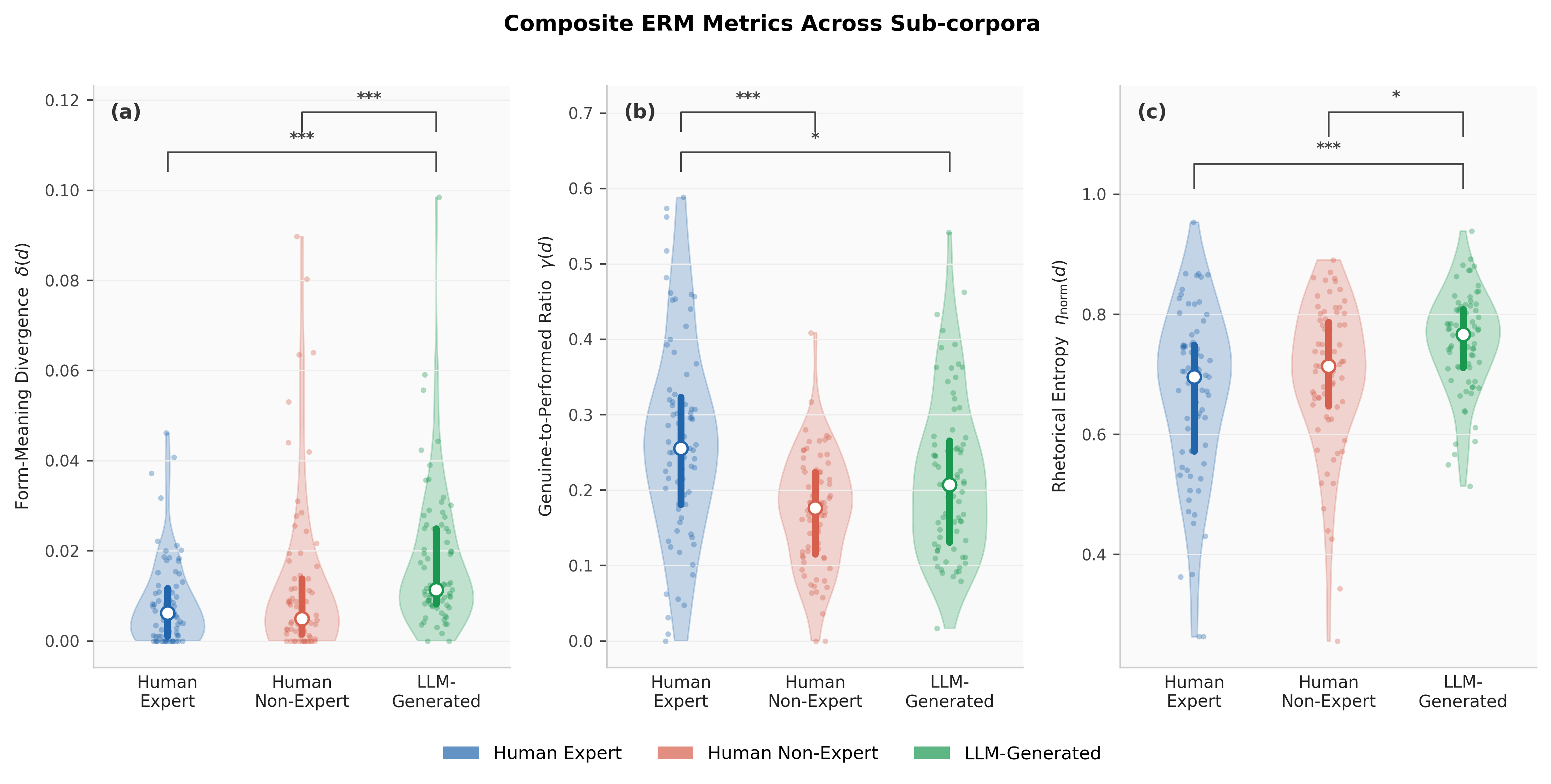}
\caption{Distribution of the three composite ERM metrics across
sub-corpora. Each violin shows the full density estimate; the
internal bar spans the interquartile range; the white dot marks
the median; jittered points show individual documents.
Significance brackets: $^{*}p{<}.05$, $^{***}p{<}.001$.
(a)~FMD $\delta(d)$. (b)~GPR $\gamma(d)$.
(c)~RDDE $\eta_{\mathrm{norm}}(d)$.}
\label{fig:composite_metrics}
\end{figure}

FMD follows a monotonic ordering across sub-corpora (Figure \ref{fig:composite_metrics}(a)). The HE texts score lowest ($\bar{\delta} = 0.009$), HN intermediate ($\bar{\delta} = 0.012$), and LG highest ($\bar{\delta} = 0.017$). Both human groups differ significantly from LG ($p < 0.001$; $\Delta = 0.68$
for HE vs LG, $\Delta = 0.30$ for HN vs LG), while the difference
between the two human groups does not reach significance as the gap is small relative to within-group variance. The dominant effect is the elevation of LLM output beyond both human groups. LLM texts combine elevated rhetorical intensity $\rho(d)$ while genuine epistemic grounding in the denominator of (\ref{eq:fmd}) does not proportionally compensate. In Gricean terms, their assertoric weight consistently exceeds the epistemic position they report.

\subsubsection{Genuine-to-Performed Epistemic Ratio}
\label{sec:results_gpr}

GPR shows an interesting three-way ordering (Figure \ref{fig:composite_metrics}(b)), with 
HE highest ($\bar{\gamma} = 0.267$), LG intermediate ($\bar{\gamma} = 0.217$), and HN lowest ($\bar{\gamma} = 0.172$) ($p < 0.001$). The largest pairwise contrast is between HE and HN texts ($\Delta = 0.89$, $p < 0.001$), reflecting expert advantage in grounding claims in identifiable evidential bases such as extensively higher modal auxiliary counts, adverbials, and syntactic
restrictors. LLM texts sit above non-experts but below experts. The LG vs HN comparison does not reach significance ($p = 0.055$), which is consistent with the prompt instruction to engage honestly with uncertainty.

\subsubsection{Rhetorical Device Distribution Entropy}
\label{sec:results_rdde}

LLM-generated texts show (Figure \ref{fig:composite_metrics}(c)) the highest and most consistent RDDE
($\bar{\eta}_{\text{norm}} = 0.753, \sigma_{\eta_\text{norm}}=0.083$) compared to HE ($\bar{\eta}_{\eta_\text{norm}} = 0.666, \sigma_\text{norm}=0.143$; $\Delta = 0.74$, $p < 0.001$) and HN ($\bar{\eta}_{\text{norm}} = 0.697$; $\Delta = 0.53$, $p = 0.011$) texts. The human groups do not differ significantly from each other ($p = 0.573$). Higher entropy indicates more uniform device distribution across the document consistent with stylistically enforcing habit rather than argumentative occasion. The lower LG variance, on the other hand,  points to template-driven generation producing cross-document consistency absent in human writing.

\subsection{Discourse-Level Structural Markers}
\label{sec:results_level3}

\begin{figure}[htbp]
\centering
\includegraphics[width=\textwidth]{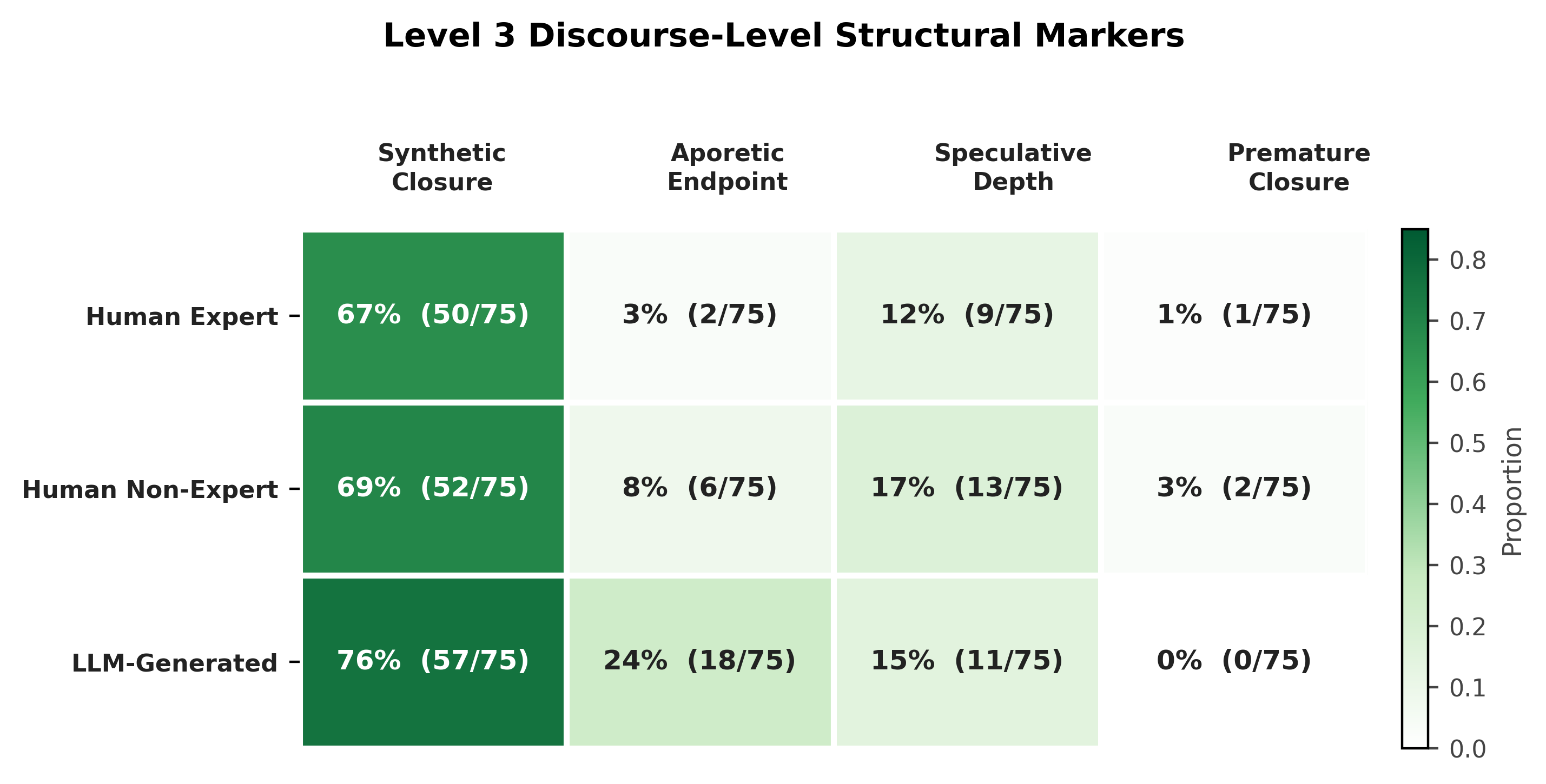}
\caption{Corpus-level proportions of Level~3 discourse markers
per sub-corpus. Cell values show the percentage and raw count
of documents annotated with each marker. Color intensity
encodes proportion. Column headers show significance of
chi-squared tests.}
\label{fig:level3_heatmap}
\end{figure}

Out of corpus-level proportions for the four Level~3 markers, only the aporetic endpoint reaches significance (Figure~\ref{fig:level3_heatmap}). Synthetic closure dominates across all three sub-corpora (HE: 66.7\%, HNE: 69.3\%, LG: 76.0\%, $p = 0.433$), consistent with the argumentative essay genre tending toward resolution regardless of author type. Premature closure and speculative depth show no significant differences.

Aporetic endpoint is elevated in LG texts (24.0\%) relative to HN (8.0\%) and HE (2.7\%) texts ($p < 0.001$). This counters the
intution that LLMs default to synthetic closure. A plausible interpretation being that the aporetic form may be a
learnable surface pattern, i.e., open-ended phrases (such as \textit{remains an open question}) deployed reflexively without genuine evidential underdetermination, constituting a discourse-level instance of performed hesitancy.

\subsection{Model-Level Variation Within the LLM Sub-corpus}
\label{sec:results_models}

\begin{figure}[htbp]
\centering
\includegraphics[width=\textwidth]{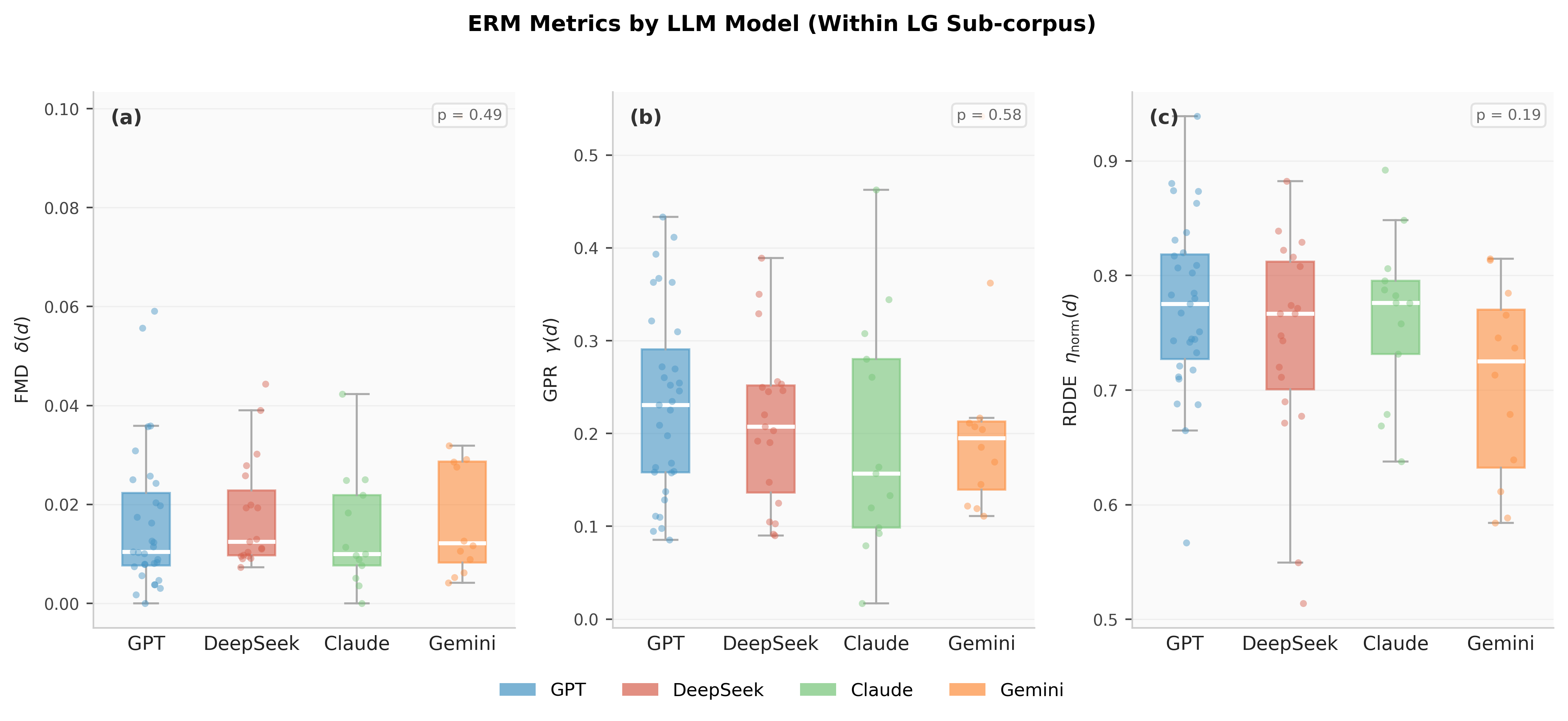}
\caption{Distribution of composite ERM metrics within the
LLM-generated sub-corpus by model. Box spans the interquartile
range; white line marks the median; jittered points show
individual documents. Kruskal-Wallis $p$-values are displayed
in the top-right corner of each panel.
(a)~FMD $\delta(d)$. (b)~GPR $\gamma(d)$.
(c)~RDDE $\eta_{\mathrm{norm}}(d)$.}
\label{fig:model_comparison}
\end{figure}

Figure~\ref{fig:model_comparison} presents ERM metric distributions
by model within the LLM sub-corpus ($n_\text{gpt} = 31$; $n_\text{deepseek} = 19$; $n_\text{claude} = 13$; $n_\text{gemini} = 12$). No significant differences are observed across models on any composite metric.

This null result is itself a substantive finding. The
epistemic-rhetorical miscalibration pattern is not an artifact of any single model's generation strategy but appears to be a systematic property of the LLM generation paradigm as a whole. The ERM framework, therefore, is potentially robust to model variations.

\subsection{Discussion, Limitations and Future Directions}
\label{sec:discussion}

The findings converge on a coherent characterisation of LLM
epistemic-rhetorical miscalibration operating across multiple
linguistic levels. At the device level, LLM texts favour tricolon, deployable without argumentative occasion, while suppressing
erotema, which requires one. At the epistemic level, the
miscalibration is specifically a \textit{performed hesitancy
excess}. LLMs produce genuine markers at a reasonable rate but
performed markers at twice the human rate. Additionally, the two human groups remain indistinguishable on performed hesitancy, confirming this as an LLM-specific signature. At the discourse level, elevated FMD and uniform RDDE jointly indicate that rhetorical elaboration runs independently of argumentative structure. 

Taken together, these patterns are consistent with the theoretical
intuitions supplied by Gricean pragmatics, Relevance Theory, and
Brandomian inferentialism. LLM texts deploy assertoric weight
that exceeds their epistemic position, impose processing costs
without proportionate epistemic returns, and undertake discursive
commitments that their visible inferential entitlements do not
adequately license.

However, there are several noteworthy limitations. The corpus covers English argumentative prose only. Generalisation to other
languages, genres, and stylistic domains remains untested. LLM annotation
was used for scalability without formal, all-encompassing, inter-annotator
reliability established against human coders, considering it beyond the scope of a philosophically inspired, exploratory framework. The prompt instruction to engage honestly with uncertainty likely suppressed performed hesitancy in the LG sub-corpus, therefore, the reported FMD values may still underestimate miscalibration in unconstrained LLM output normally seen in digital content all over internet. Specific markers with extended LG sub-corpus with variable model representation may give further insights into epistemic alignment of different model variants. 

Finally, the ERM taxonomy is designed to measure empirically falsifiable, optimizable and programmatically discernible surface-linguistic markers. The underlying, more complex epistemic states remain a subject of close-reading augmented, qualitative linguistic enquiry.

\section{Conclusion}
\label{sec:conclusion}
This study introduced a framework for quantifying
epistemic-rhetorical miscalibration in large language models through annotation of rhetorical devices, epistemic stance markers, and discourse-level argumentative structure. Applied to a medium sized  corpus across three author types, the
framework identifies a consistent and model-agnostic LLM
signature, including elevated tricolon density, suppressed erotema, and performed hesitancy at twice the human rate, replicated across four frontier models and grounded in Gricean pragmatics, Relevance Theory, and Brandomian inferentialism.

The ERM pipeline is fully automatable and executable by any capable language model against any argumentative text. This makes it deployable as a lightweight miscalibration screening tool for better epistemic alignment of target texts. Significant individual markers such as tricolon density and performed
hesitancy density could also serve as features in LLM-generated text detection pipelines, complementing existing stylometric approaches with theoretically motivated epistemic signals. FMD and GPR further operationalize properties not captured by standard benchmarks, measuring proportionality of assertoric
weight to epistemic position and calibration of rhetorical form to evidential grounding.
\clearpage
\appendix
\section{ERM Taxonomy}
\label{app:erm}
\begin{table}[htbp]
\centering
\caption{\textbf{Level 1a -- Sentence-level Rhetorical Devices.}
Devices that operate within a single sentence to modulate
presentational intensity through syntactic and phonological structure.}
\label{tab:erm_1a}
\vspace{2pt}
\small
\setlength{\tabcolsep}{5pt}
\renewcommand{\arraystretch}{1.5}

\begin{tabularx}{\textwidth}{@{}
  >{\RaggedRight\arraybackslash}p{2cm}
  >{\RaggedRight\arraybackslash}p{6cm}
  >{\RaggedRight\arraybackslash}X@{}}

\toprule
\textbf{Device} & \textbf{Definition} & \textbf{Example} \\
\midrule

Tricolon
  & Three parallel syntactic units in succession, producing
    rhetorical completeness through numerical symmetry.
  & \textit{``The hypothesis is parsimonious, falsifiable,
    and empirically grounded.''} \\

\midrule

Anaphora
  & Identical word or phrase repeated at the opening of
    consecutive clauses, generating cumulative emphasis.
  & \textit{``Matter has mass. Matter has extension.
    Matter has inertia.''} \\

\midrule

Chiasmus
  & AB--BA inversion of grammatical or semantic elements
    across two successive clauses, foregrounding opposition
    or resolution.
  & \textit{``We do not ask what language knows; we ask
    what language does.''} \\

\midrule

Erotema
  & A rhetorical question that asserts rather than enquires,
    presupposing a shared answer and enlisting reader assent.
  & \textit{``Can a calculus that has never been applied
    be said to have mathematical content?''} \\

\midrule

Sententia
  & A self-contained aphoristic statement closing an argument
    with general, present-tense propositional force; free of
    first- and second-person deixis.
  & \textit{``Form without warrant is persuasion without
    truth.''} \\

\bottomrule
\end{tabularx}
\end{table}

\begin{table}[htbp]
\centering
\caption{\textbf{Level 1b -- Argument-level Rhetorical Devices.}
Devices that organise the internal structure of an argument
across multiple sentences or sub-claims.}
\label{tab:erm_1b}
\vspace{2pt}
\small
\setlength{\tabcolsep}{5pt}
\renewcommand{\arraystretch}{1.5}

\begin{tabularx}{\textwidth}{@{}
  >{\RaggedRight\arraybackslash}p{2cm}
  >{\RaggedRight\arraybackslash}p{6cm}
  >{\RaggedRight\arraybackslash}X@{}}

\toprule
\textbf{Device} & \textbf{Definition} & \textbf{Example} \\
\midrule

Correctio
  & A mid-argument self-correction that retracts or intensifies
    a prior formulation via explicit metalinguistic markers,
    performing deliberateness rather than genuine revision.
  & \textit{``The data support the hypothesis---or rather,
    are consistent with it under one interpretation.''} \\

\midrule

Enumeratio
  & Structured listing of $\geq 3$ supporting points or
    sub-claims introduced by ordinal cues, creating the
    rhetorical impression of comprehensive coverage.
  & \textit{``Three factors drive the result: sampling
    variance, measurement error, and model
    misspecification.''} \\

\midrule

Auxesis
  & Successive clauses or phrases arranged in strictly
    ascending semantic or affective intensity, producing
    a climactic crescendo effect.
  & \textit{``The discrepancy is notable, statistically
    significant, and theoretically fatal to the
    standard account.''} \\

\bottomrule
\end{tabularx}
\end{table}

\begin{table}[htbp]
\centering
\caption{\textbf{Level 1c -- Narrative-level Rhetorical Devices.}
Devices that operate across the full span of a discourse unit,
shaping its overall trajectory and producing large-scale
rhetorical effects such as surprise or discovery.}
\label{tab:erm_1c}
\vspace{2pt}
\small
\setlength{\tabcolsep}{5pt}
\renewcommand{\arraystretch}{1.5}

\begin{tabularx}{\textwidth}{@{}
  >{\RaggedRight\arraybackslash}p{2cm}
  >{\RaggedRight\arraybackslash}p{6cm}
  >{\RaggedRight\arraybackslash}X@{}}

\toprule
\textbf{Device} & \textbf{Definition} & \textbf{Example} \\
\midrule

Peripeteia
  & A sudden reversal of the argument's established evaluative
    or epistemic direction, in which an apparent trajectory
    is abruptly inverted at the discourse level.
  & \textit{``All measurements confirmed the classical
    prediction. Yet the 1919 eclipse data required a
    fundamentally different theory to explain them.''} \\

\midrule

Expectation-\newline Violation-\newline Resolution (EVR)
  & A three-phase arc in which (i) an expectation is
    established, (ii) explicitly violated, and (iii) a
    resolution or reframing is offered; produces the
    rhetorical effect of discovery or insight.
  & \textit{``For decades, the prevailing assumption held that stomach ulcers were caused by stress and excess acid, and treatment focused accordingly on antacids and lifestyle changes [E]. In 1984, Marshall and Warren demonstrated that the majority of cases were caused by a bacterial infection, \textit{Helicobacter pylori} [V]. This reframed ulcers as an infectious disease, making them curable with a standard course of antibiotics [R].''} \\

\bottomrule
\end{tabularx}
\end{table}

\begin{table}[htbp]
\centering
\caption{\textbf{Level 2a -- Genuine Epistemic Stance Markers.}
Lexical and syntactic devices that ground a speaker's commitment to a claim in an identifiable evidential base or calibrated modal force.}
\label{tab:erm_2a}
\vspace{2pt}
\small
\setlength{\tabcolsep}{7pt}
\renewcommand{\arraystretch}{1.55}

\begin{tabularx}{\textwidth}{@{}
  >{\RaggedRight\arraybackslash}p{3cm}
  >{\RaggedRight\arraybackslash}p{6.2cm}
  >{\RaggedRight\arraybackslash}X@{}}

\toprule
\textbf{Marker} &
\textbf{Definition} &
\textbf{Example} \\
\midrule

Modal auxiliaries
  & Modals expressing epistemic necessity (\textit{must}) or
    possibility (\textit{might}, \textit{may}, \textit{could})
    relative to an evidential base. \textit{Ought} and
    \textit{should} signal weak inferential necessity.
  & \textit{``Given the rate of polar ice loss, sea levels
    \textbf{must} be rising faster than nineteenth-century
    models predicted.''} \\

\midrule

Adverbial expressions
  & Sentence adverbials calibrating speaker commitment across
    a gradient from certainty (\textit{certainly},
    \textit{clearly}) through probability (\textit{probably},
    \textit{likely}) and possibility (\textit{possibly},
    \textit{perhaps}) to source attribution
    (\textit{apparently}, \textit{allegedly}).
  & \textit{``The replication failure \textbf{probably}
    reflects sampling variance rather than a genuine
    reversal of the effect.''} \\

\midrule

Syntactic restrictor phrases
  & Constructions that restrict the evidential base of a
    modal or judgement: conditional \textit{if}-clauses,
    parenthetical source attributions, and restrictor
    prepositional phrases (\textit{judging by X},
    \textit{according to Y}, \textit{as far as Z knows},
    \textit{given what we know}).
  & \textit{``\textbf{Judging by the biopsy results}, the
    tumour \textit{may} respond to targeted therapy.''} \\

\midrule

Evidential markers
  & Markers specifying the type of evidence underlying a
    claim: \textbf{direct} (sensory access: \textit{we
    observed}, \textit{we measured}); \textbf{indirect
    reported} (cited source: \textit{according to},
    reporting verbs); \textbf{indirect inferential}
    (reasoning from evidence: \textit{suggests},
    \textit{implies}, \textit{indicates}).
  & \textit{``\textbf{We measured} a 0.3~mg deviation
    across all trials.''} (direct)\quad
    \textit{``The stratigraphic record \textbf{suggests}
    deposition under anaerobic conditions.''} (inferential) \\

\bottomrule
\end{tabularx}
\end{table}

\begin{table}[htbp]
\centering
\caption{\textbf{Level 2b -- Performed Epistemic Stance Markers.}
Surface devices that adopt the register of uncertainty or reflexivity without providing the evidential grounding, modal calibration, or
restrictor structure that genuine epistemic marking presupposes.}
\label{tab:erm_2b}
\vspace{2pt}
\small
\setlength{\tabcolsep}{7pt}
\renewcommand{\arraystretch}{1.55}

\begin{tabularx}{\textwidth}{@{}
  >{\RaggedRight\arraybackslash}p{2.7cm}
  >{\RaggedRight\arraybackslash}p{6.2cm}
  >{\RaggedRight\arraybackslash}X@{}}

\toprule
\textbf{Marker} &
\textbf{Definition} &
\textbf{Example} \\
\midrule

Complexity tokens
  & Fixed phrasal chunks that invoke irreducible complexity
    without identifying a source of uncertainty, restricting
    an evidential base, or calibrating modal force. They
    signal that a claim is difficult without specifying
    \textit{why} or \textit{for whom}, producing the
    surface appearance of epistemic humility without its
    substance.
  & \textit{``The relationship between consciousness and
    neural activity is \textbf{deeply nuanced} and admits
    \textbf{no easy answers}.''} \\

\midrule

Meta-epistemic gestures
  & Utterances that foreground conditionality or perspective relativity as a performance of reflexivity, without specifying the condition or perspective. The hedging form is present but the epistemic content it presupposes is absent. 
  & \textit{``\textbf{It depends} on how one defines
    causation.''}\quad
    \textit{``\textbf{From a certain perspective}, free
    will and determinism are compatible.''} \\

\bottomrule
\end{tabularx}
\end{table}

\begin{table}[htbp]
\centering
\caption{\textbf{Level 3 -- Discourse-level Argumentative Structure.} Global properties of the argument's endpoint and overall inferential trajectory, identified at the level of the
whole text rather than the sentence or clause.}
\label{tab:erm_3}
\vspace{2pt}
\small
\setlength{\tabcolsep}{7pt}
\renewcommand{\arraystretch}{1.55}

\begin{tabularx}{\textwidth}{@{}
  >{\RaggedRight\arraybackslash}p{2cm}
  >{\RaggedRight\arraybackslash}p{6cm}
  >{\RaggedRight\arraybackslash}X@{}}

\toprule
\textbf{Marker} &
\textbf{Definition} &
\textbf{Example} \\
\midrule

Aporetic Endpoint
  & The argument terminates in explicitly sustained uncertainty, withholding resolution on the grounds that the available evidence underdetermines a conclusion. Calibrated when the evidential base warrants suspension of judgement.
  & \textit{``Whether dark matter consists of weakly interacting massive particles or axions remains an open question that current detection sensitivities cannot resolve.''} \\

\midrule

Synthetic Closure
  & The argument arrives at a positive, integrated conclusion that synthesises prior competing considerations into a unified claim, signalled by conclusive discourse connectives (\textit{therefore}, \textit{thus}, \textit{taken together}).
  & \textit{``Taken together, the fossil record, the genetic data, and the biogeographic distribution \textbf{confirm} that the two populations diverged no later than the Pleistocene.''} \\

\midrule

Premature Closure
  & A closure move in which resolution is asserted before the inferential structure of the argument warrants it. Typically signalled by explicit certainty tokens (\textit{it is clear that}, \textit{obviously}, \textit{without question}) that short-circuit rather than conclude the argument.
  & \textit{``While these questions are admittedly complex, \textbf{it is clear} that the hard problem of consciousness is a pseudoproblem that dissolves under a sufficiently rigorous functionalist     analysis.''} \\

\midrule

Speculative Depth
  & A sustained chain of conditional or hypothetical reasoning that extends inferentially beyond the argument's established evidential base, characterised     by nested if-then structures or multiple stacked modal scopes maintained across several sentences.
  & \textit{``If LLMs encode syntactic structure, and if that encoding extends to semantic compositionality, then such systems \textit{might} possess a rudimentary form of understanding -- though what understanding would mean here remains entirely unclear.''} \\

\bottomrule
\end{tabularx}
\end{table}

\FloatBarrier
\clearpage

\bibliographystyle{unsrtnat}
\bibliography{references}  

\end{document}